%% file: main.tex
\documentclass[10pt,twocolumn,letterpaper]{article}

\usepackage[pagenumbers]{iccv} 

\input{preamble}

\definecolor{iccvblue}{rgb}{0.21,0.49,0.74}
\usepackage[pagebackref,breaklinks,colorlinks,allcolors=iccvblue]{hyperref}

\def\paperID{11908} 
\def\confName{ICCV}
\def\confYear{2025}

\title{Representation Shift: Unifying Token Compression with FlashAttention}

\newcommand*\samethanks[1][\value{footnote}]{\footnotemark[#1]}
\author{\textbf{Joonmyung Choi}$^1$\thanks{Equal contribution.} \hspace{0.4cm} \textbf{Sanghyeok Lee}$^1$\samethanks \hspace{0.4cm}  \textbf{Byungoh Ko}$^1$ \\\textbf{Eunseo Kim}$^1$ \hspace{0.4cm}  \textbf{Jihyung Kil}$^2$ \hspace{0.4cm} \textbf{Hyunwoo J. Kim}$^3$\thanks{Corresponding author.} 
\\
$^1$Korea University \hspace{0.4cm} $^2$Adobe Research \hspace{0.4cm}
$^3$KAIST\\
{\tt\small \{pizard, cat0626, ko990128, pingdoll3110\}@korea.ac.kr}\\
{\tt\small jkil@adobe.com \hspace{0.4cm} hyunwoojkim@kaist.ac.kr}
}

\begin{document}

\maketitle


 
\input{sec/0_abstract}
\input{sec/1_intro}
\input{sec/2_rel}
\input{sec/3_method}

\input{sec/4_exp}
\input{sec/5_con}

{
    \small
    \bibliographystyle{ieeenat_fullname}
    \bibliography{main}
}

\end{document}

%% file: preamble.tex
\usepackage{multirow}

%% file: sec/0_abstract.tex
\begin{abstract}
Transformers have demonstrated remarkable success across vision, language, and video.
Yet, increasing task complexity has led to larger models and more tokens, raising the quadratic cost of self-attention and the overhead of GPU memory access.
To reduce the computation cost of self-attention, prior work has proposed token compression techniques that drop redundant or less informative tokens.
Meanwhile, fused attention kernels such as FlashAttention have been developed to alleviate memory overhead by avoiding attention map construction and its associated I/O to HBM.
This, however, makes it incompatible with most training-free token compression methods, which rely on attention maps to determine token importance.
Here, we propose Representation Shift, a training-free, model-agnostic metric that measures the degree of change in each token's representation.
This seamlessly integrates token compression with FlashAttention, without attention maps or retraining.
Our method further generalizes beyond Transformers to CNNs and state space models.
Extensive experiments show that Representation Shift enables effective token compression compatible with FlashAttention, yielding significant speedups of up to 5.5× and 4.4× in video-text retrieval and video QA, respectively.
Code is available at \url{https://github.com/mlvlab/Representation-Shift}.
\end{abstract}

%% file: sec/1_intro.tex
\input{figure/fig_intro}
\section{Introduction}
\label{sec:intro}
Transformers, initially proposed for natural language processing (NLP)~\cite{vaswani2017attention}, have become a prominent architecture in the vision domain.
Following the pioneering work ViTs~\cite{dosovitskiy2020image}, numerous subsequent studies have extended Transformers to various vision tasks, \eg, image classification~\cite{liu2021swin,chu2021twins,touvron2021going,touvron2021training,wang2022pvt,dosovitskiy2020image}, object detection~\cite{carion2020end,wang2021anchor,wang2021pyramid,zhu2020deformable, zhang2022dino,kim2024groupwise}, segmentation~\cite{zheng2021rethinking,cheng2021per,strudel2021segmenter}, and video understanding~\cite{li2023unmasked, wang2022internvideo,wang2024internvideo2, tong2022videomae,wang2021end,ko2023meltr,ko2022video,park2025deepvideo,lee2025vidchain}.
While these works have proven to be effective, the quadratic complexity of the self-attention mechanism remains a critical bottleneck, limiting the scalability of Transformer based architectures.\\
To address this problem, a wide range of approaches have been proposed to accelerate Transformers across various domains, such as vision and natural language processing (NLP).
Early works tackled the computational burden by proposing sparse attention mechanisms~\cite{beltagy2020longformer, kitaev2020reformer, xiong2021nystromformer, wang2020linformer} and architectural modifications~\cite{liu2021swin, chu2021twins, roy2021efficient, zaheer2020big, wang2022pvt, tu2022maxvit} to approximate self-attention, such as low-rank approximations and sparse attention patterns.
However, these methods often introduce structural deviations from the original Transformer architecture, making them incompatible with widely adopted pretrained models.
As a result, vanilla Transformers~\cite{dosovitskiy2020image} remain the dominant choice in practice, supported by the widespread availability of pre-trained models across a variety of domains.
Here, one promising approach to accelerate pre-trained vanilla Transformers is FlashAttention~\cite{dao2022flashattention}, which optimizes GPU memory access of self-attention while maintaining the original formulation.
While FlashAttention preliminarily focused on the long sequences of LLM, it also demonstrates substantial acceleration with Vision Transformers as in recent works~\cite{wang2024internvideo2,chen2024internvl,alif2025yolov12,oquab2023dinov2,wang2023videocomposer}.\\
Another line of work in accelerating Vision Transformers is token compression~\cite{liang2022not, long2023beyond, wang2024zero, choi2024vid, bolya2022token, yuan2024efficient, lee2024multi, pan2021ia, kim2022learned, meng2022adavit, rao2021dynamicvit, yin2022vit}, which reduces computational cost by pruning or merging tokens.
Since determining which tokens to retain is crucial, previous works incorporate token importance measurement as a fundamental step.
Some approaches~\cite{meng2022adavit,rao2021dynamicvit,yin2022vit} introduce additional learnable networks to predict token importance, and other works~\cite{graham2021levit,long2023beyond,lee2024multi,wang2024zero,choi2024vid} employ attention-based heuristics as a surrogate for token importance.
Although these works have shown promising acceleration on Vision Transformers, methods that employ learnable networks necessitate extra training, making them infeasible in a training-free manner.
Also, attention-based scoring methods limit their use when the attention map is unavailable (e.g., FlashAttention, CNN).
While FlashAttention alone provides substantial acceleration, achieving a $1.5\times$ speedup on DeiT-S and $2.7\times$ speedup on UMT-B, existing token pruning methods fail to further improve efficiency in a training-free setting due to their reliance on learnable modules or attention maps.\\
To address this, we propose a token importance criterion that is training-free and model-agnostic, based on representation shift, which quantifies the change in token embeddings before and after the layer(\Cref{fig:main}).
This simple but effective approach successfully captures the amount of information amplified by any operation, \textit{e.g.}, FFN, Attention, and Convolutions.
By leveraging representation shift as an importance metric, our method effectively identifies and removes redundant tokens. 
Since our method is not dependent on attention mechanisms, it generalizes beyond Transformers to architectures like CNNs~\cite{he2016deep,liu2022convnet,woo2023convnext} and SSMs~\cite{zhu2024vision,liu2024vmamba,lee2025efficientvim}, while seamlessly integrating with fused kernel operations such as FlashAttention for efficient inference.
Experimental results show that our method outperforms existing attention-based token importance methods in both accuracy and efficiency on vanilla Transformers.
Specifically, we achieve impressive throughput improvements of about 5.5$\times$ speedup with UMT~\cite{li2023unmasked} on multiple video-text retrieval benchmarks.
Moreover, unlike prior attention-dependent methods, our approach additionally generalizes to previously unsupported architectures such as CNNs and state space models.
In sum, our key contributions are as follows:
\begin{itemize}
    \item We propose a novel approach for estimating token importance, called \textbf{representation shift}, which directly captures the amount of information amplified by each operation. 
    This model-agnostic importance score can be computed in a training-free manner with negligible overhead.
    \item To the best of our knowledge, this is the first token reduction method applicable to both FlashAttention and CNNs.
    \item Through extensive experiments on video and image understanding tasks, we demonstrate that combining FlashAttention with our representation shift–based token pruning yields notable inference speedups.
\end{itemize}

%% file: figure/fig_intro.tex
\begin{figure}
     \centering
     \captionsetup{type=figure}
     \includegraphics[width=0.42\textwidth]{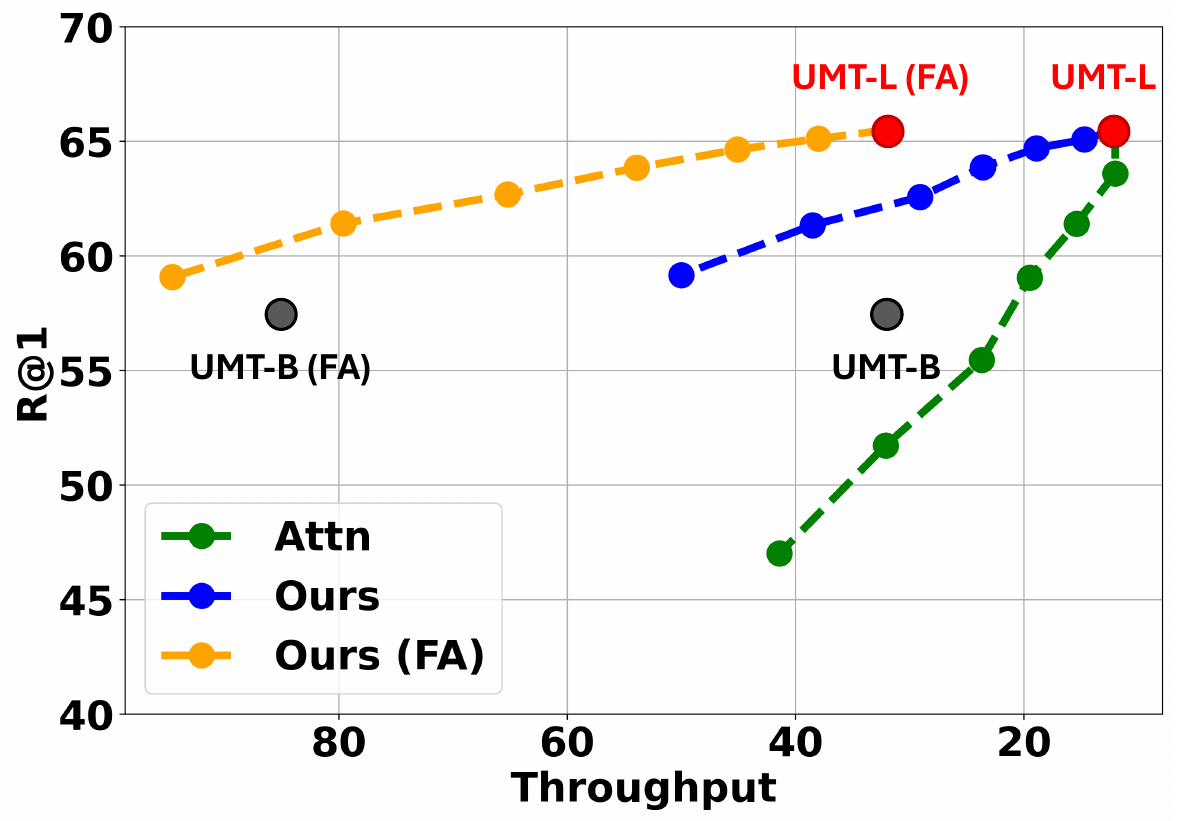}
     \vspace{-10pt}
     \captionof{figure}{Comparison of importance metrics for token pruning (average over 7 video-text retrieval benchmarks in \Cref{table:retrieval}). Pruning with a conventional attention-based score (Attn) yields poor speed-accuracy trade-offs on UMT-L and is incompatible with FlashAttention (FA). In contrast, our proposed representation shift accelerates both vanilla UMT-L and UMT-L with FlashAttention, achieving superior trade-offs compared to downscaling to UMT-B and attention-based scores.}
     \label{fig:intro}
     \vspace{-20pt}
\end{figure}

%% file: sec/2_rel.tex
\input{figure/fig_main}
\input{figure/fig_toy}
\section{Related works}
\label{sec:2}
\noindent\textbf{Efficient Vision Transformers.}
Built with ViTs~\cite{dosovitskiy2020image}, self-attention~\cite{vaswani2017attention} are introduced to handle various vision tasks.
Following works~\cite{liu2021efficient,yuan2021tokens}, such as DeiT~\cite{touvron2021training}, further improve data efficiency of Vision Transformers.
However, despite the competitive performance, the quadratic cost of self-attention with respect to the number of tokens remains the major bottleneck.
To address this issue, earlier works~\cite{choromanski2020rethinking,kitaev2020reformer,wang2020linformer,xiong2021nystromformer,katharopoulos2020transformers,qin2022cosformer} have tried to find an efficient approximation of self-attention.
For instance, Reformer~\cite{kitaev2020reformer} achieves the $O(N\log N)$ complexity with a hashing function, and Linformer~\cite{wang2020linformer} approximates the self-attention via a low-rank matrix, resulting in the linear cost of $O(N)$.
Nystromformer~\cite{xiong2021nystromformer} and performer~\cite{choromanski2020rethinking} also present the linear approximation of the self-attention.
In parallel, several works~\cite{beltagy2020longformer,zaheer2020big,chen2021scatterbrain,roy2021efficient} have focused on sparsifying the attention map to lessen complexity.
Similarly, recent vision transformers~\cite{liu2021swin,chu2021twins,wang2022pvt,wang2021pyramid,tu2022maxvit} reduce the number of key and value tokens. PVT~\cite{wang2022pvt,wang2021pyramid} introduce spatial-reduction attention that downsamples the key and value tokens before attention, and Swin~\cite{liu2021swin}, Twins~\cite{chu2021twins}, and MaxViT~\cite{tu2022maxvit} also apply local attention to reduce the reference tokens.
Also, for the deployment in edge-devices, a line of work~\cite{cai2022efficientvit,liu2023efficientvit,yun2024shvit,graham2021levit, mehta2021mobilevit} has been proposed. More recently, with the aim to reduce the latency by memory-bound operation, FlashAttention~\cite{dao2022flashattention} conducts attention calculation within fast SRAM minimizing the memory access to slow HBM.
In this work, we aim to further boost the FlashAttention with token compression.

\noindent\textbf{Token Compression.}
Since the cost heavily relies on the number of tokens, recent works~\cite{liang2022not, yin2022vit,rao2021dynamicvit,meng2022adavit,bolya2022token, long2023beyond, choi2024vid, kim2022learned,pan2021ia,yuan2024efficient,wang2024zero} explicitly focus on compressing the token.
To preserve the core information of an image after compressing tokens, they generally prune or merge unimportant tokens.
Importance estimation typically follows two major approaches.
First is the additional learnable network to predict the importance.
For instance, AdaViT~\cite{meng2022adavit} and DynamicViT~\cite{rao2021dynamicvit} introduce additional learnable decision networks to select the tokens to be compressed,
and A-ViT~\cite{yin2022vit} also needs to train additional parameters for calculating the importance of the tokens.
Second one is to utilize intermediate attention scores as a surrogate function for measuring the importance.
Specifically, EViT~\cite{liang2022not} and BAT~\cite{long2023beyond} approximate the importance of the tokens using the attention score for the class tokens, which indicate the influence of each token on the final prediction.
Zero-TPrune~\cite{wang2024zero} measures the informativeness of tokens via a ranking method with attention maps inspired by Page Rank~\cite{brin1998pagerank}.
In the video domain, vid-TLDR~\cite{choi2024vid} captures the salient regions based on the entropy of the attention scores.
While the aforementioned works have proven to be effective in compressing tokens with the affordable speed-accuracy trade-offs, they require either additional training or attention maps.
Note that FlashAttention does not provide intermediate attention scores to minimize memory access on HBM.
As a result, despite the much faster speed of FlashAttention over standard self-attention, it is not straightforward to apply previous token compression methods in a training-free manner.



%% file: figure/fig_main.tex
\begin{figure}[t!]
\centering
\includegraphics[width=0.95\linewidth]{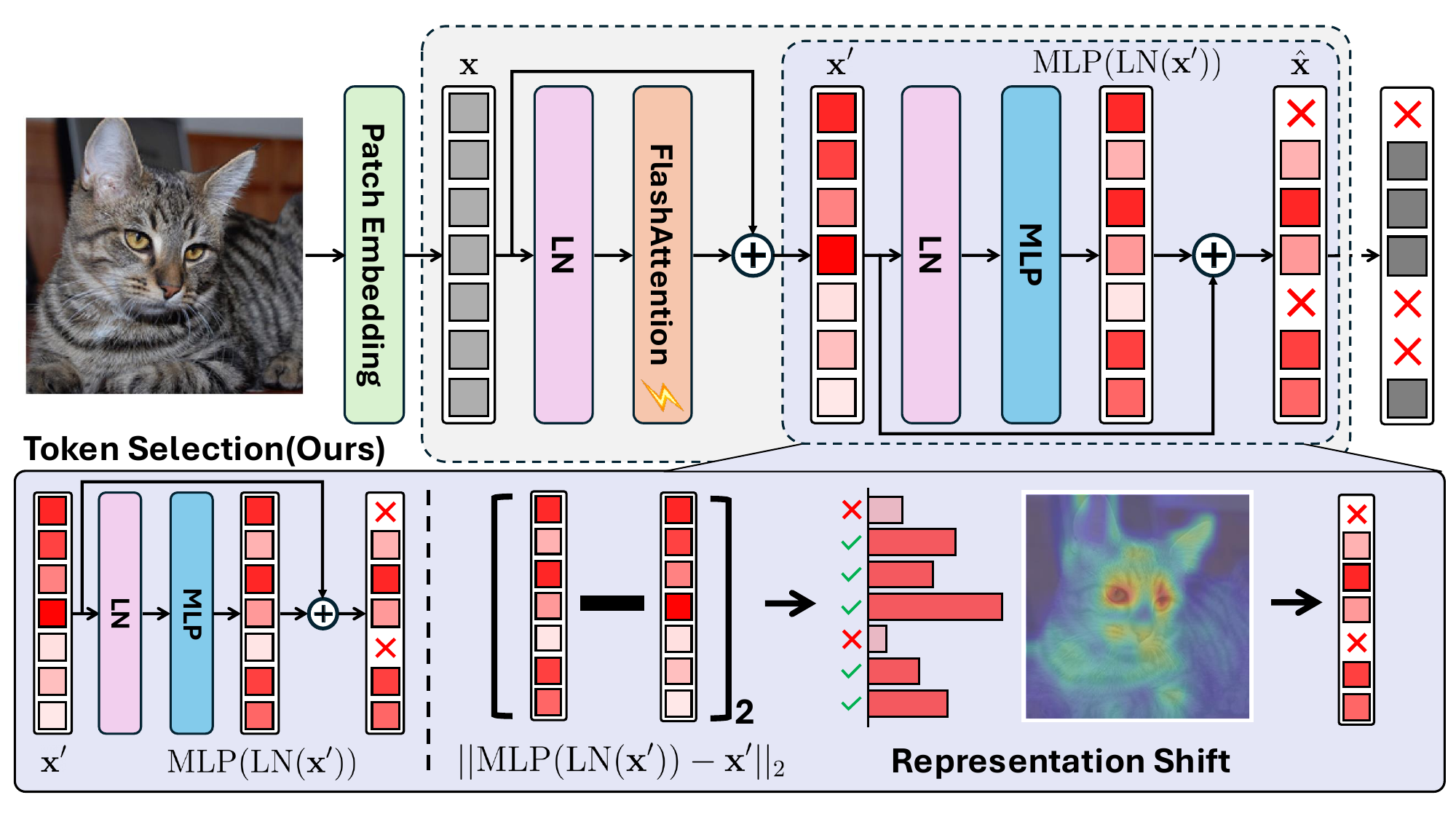}
 \caption{Illustration of representation shift for token importance. We compute the L2 distance between token representations before and after the MLP layer to quantify how much each token is emphasized by the transformation.}
 
\label{fig:main}
\vspace{-15pt}
\end{figure}

%% file: figure/fig_toy.tex
\begin{figure*}[t!]
  \centering
  \begin{minipage}[b]{0.34\textwidth}
    \centering
    \setlength{\tabcolsep}{3pt}
    \begin{tabular}{c|cc|cc}
      \toprule
      \multirow{2}{*}{Method} & \multicolumn{2}{c|}{DeiT-S~\cite{touvron2021training}} & \multicolumn{2}{c}{UMT-B~\cite{li2023unmasked}} \\ 
      & Acc-1 & Thr & R@1 & Thr  \\
      \midrule
      Self-Attention & 79.8 & 2308 & 50.0 & 32 \\ 
      FlashAttention & 79.8 & \textbf{3552} & 50.0 & \textbf{85} \\ 
      \bottomrule
    \end{tabular}
    \vspace{+10pt}
    \captionof{table}{Comparison of FlashAttention~\cite{dao2022flashattention} with standard self-attention. Throughputs are measured with NVIDIA RTX A6000. ImageNet~\cite{deng2009imagenet} and MSRVTT~\cite{xu2016msr} are used for image and video understanding, respectively.}
    \label{tab:flash}
  \end{minipage}
  \hfill
  \begin{minipage}[b]{0.63\textwidth}
    \centering
    \begin{subfigure}[b]{0.48\textwidth}
      \centering
      \includegraphics[width=\textwidth]{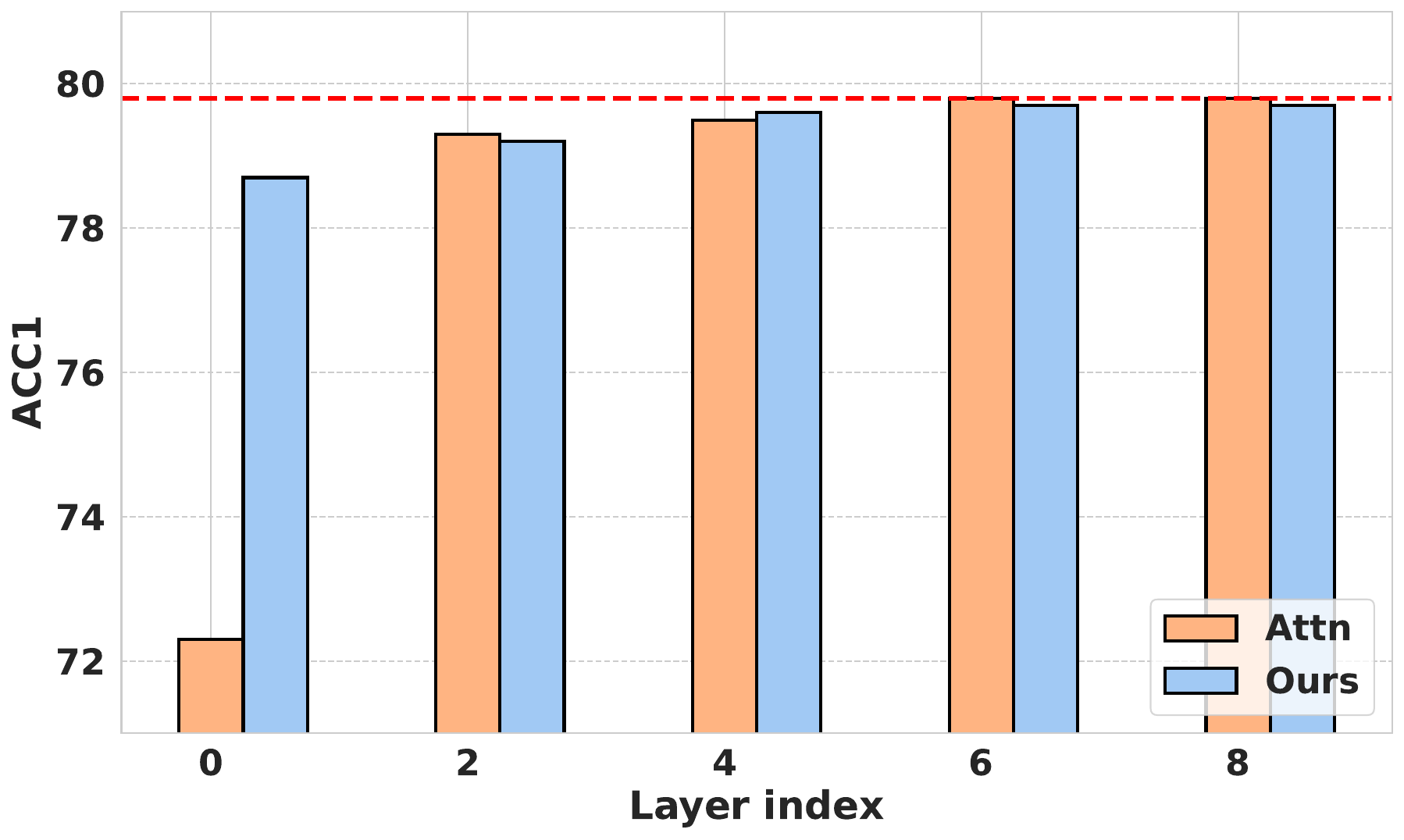}
      \caption{DeiT-S}
    \end{subfigure}
    \hfill
    \begin{subfigure}[b]{0.48\textwidth}
      \centering
      \includegraphics[width=\textwidth]{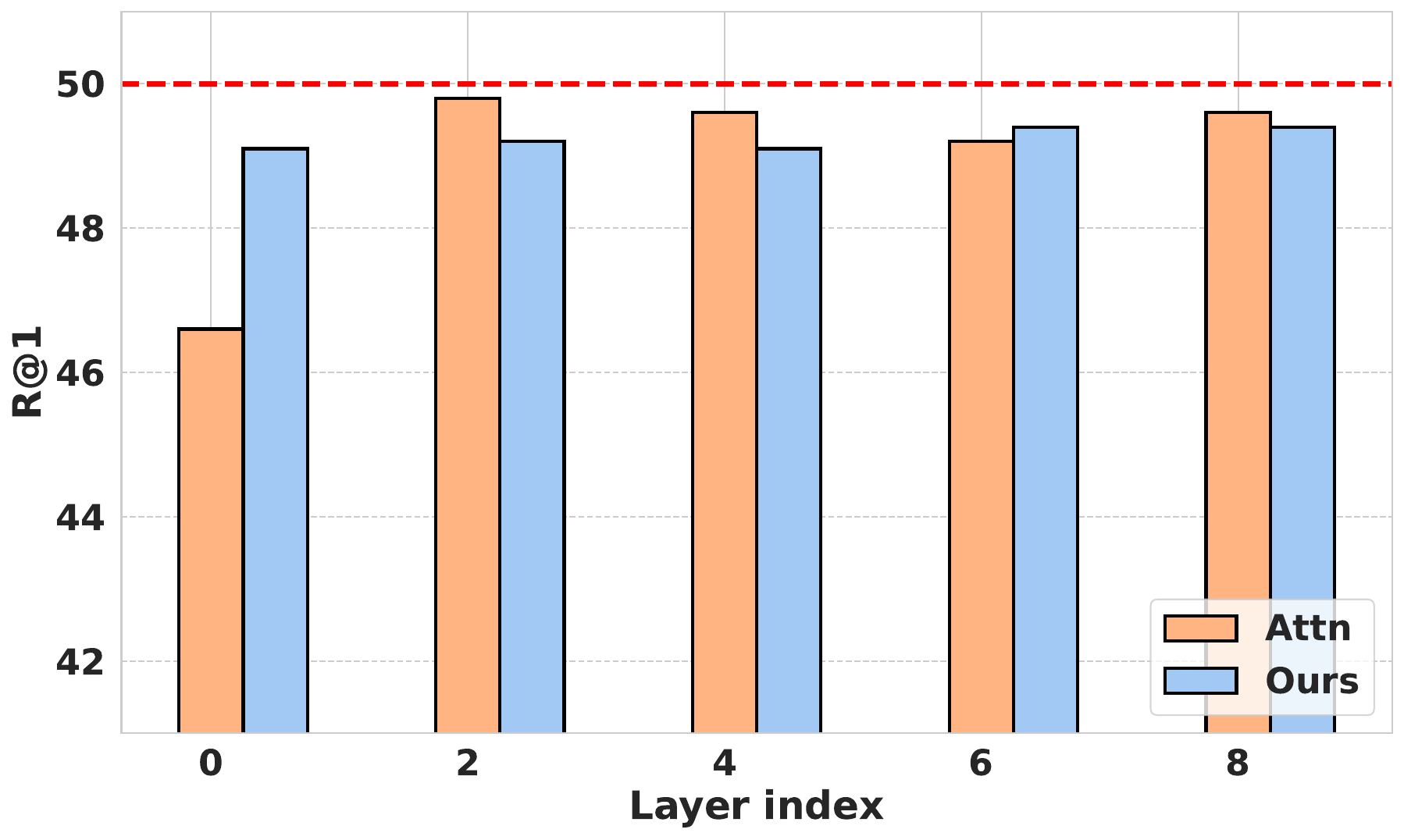}
      \caption{UMT-B}
    \end{subfigure}
    \captionof{figure}{Comparison of representation shift and attention-based scores as importance for token pruning. For DeiT~\cite{touvron2021training} and UMT~\cite{li2023unmasked}, 40 and 1100 tokens are pruned at each layer. The red line indicates baseline performance without token compression.}
    \label{fig:toy}
  \end{minipage}
\vspace{-10pt}
\end{figure*}

%% file: sec/3_method.tex
\section{Method}
\label{sec:3}
\subsection{Preliminaries}
In Vision Transformers~\cite{dosovitskiy2020image,touvron2021going,touvron2021training}, the input image is first partitioned into a set of image patches $\mathbf{x} \in \mathbb{R}^{N \times C}$, called tokens, where $N = \frac{H}{P} \times \frac{W}{P}$ is the number of tokens, $H\times W$ is the resolution of the image, and $P$ is the patch size.
This set of tokens is then processed via self-attention defined as:
\begin{equation}
    \text{SA}(\mathbf{x}) = \text{Softmax}\left(\frac{QK^\top}{\sqrt{C}}\right)V,
    \label{eq:sa}
\end{equation}
where $[Q,K,V] = \mathbf{x}W$, $W \in \mathbb{R}^{C \times 3C}$ is a learnable projection matrix.
This process incurs the quadratic cost of $O(N^2C + NC^2)$.
To mitigate this cost, recent works~\cite{liang2022not, yin2022vit,rao2021dynamicvit,choi2024vid,meng2022adavit} explicitly prune less informative tokens, resulting in a reduced token set $\tilde{\mathbf{x}} \in \mathbb{R}^{(N-r) \times C}$, where $r$ is the number of pruned tokens.\\
The importance of tokens, $s \in \mathbb{R}^{L}$, is typically estimated using the attention map, $A = \text{Softmax}\left(\frac{QK^\top}{\sqrt{C}}\right)$, which is the byproduct of self-attention. 
For example, the importance of the tokens can be defined as the attention scores relative to the class token $q_\text{cls} \in \mathbb{R}^{1 \times C}$:
\begin{equation}
    s = \text{Softmax}\left(\frac{q_\text{cls}K^\top}{\sqrt{C}}\right),
    \label{eq:cls}
\end{equation}
or as a summarized attentiveness across all query vectors:
\begin{equation}
    s = \frac{1}{N}\sum^N_i A_i,
    \label{eq:sum}
\end{equation}
where $A_i = \text{Softmax}\left(\frac{q_iK^\top}{\sqrt{C}}\right)$.
While these attention-based scores have proven effective as surrogate measures for the informativeness of the tokens, they are not applicable when the attention map is unavailable, as in the case of FlashAttention~\cite{dao2022flashattention}.
In our preliminary experiments (\Cref{tab:flash}), FlashAttention also brings substantial speedup over standard attention in Vision and Video Transformers, \textit{e.g.}, DeiT~\cite{touvron2021training} and UMT~\cite{li2023unmasked}.
Despite the promising results, we cannot further boost it with previous attention-based token compressions.
Here, we aim to develop a simple yet effective model-agnostic method for quantifying token importance in a training-free manner.

\subsection{Representation shift for token importance}
\label{sec:3.2}
In our preliminary experiments, we observed that the representation shifts of the tokens through a network layer reflect their contribution to the prediction of the model.
Here, we first define the representation shift, and then provide the quantitative and qualitative results to validate it.
Given input tokens $\mathbf{x}\in \mathbb{R}^{L \times C}$, the representation shift for importance score $s$ is defined as
\begin{equation}
s = \Delta \mathbf{x} = \mathcal{D} (F(\mathbf{x}), \mathbf{x}) ,
\end{equation}
where $F(\cdot)$ indicates the transformation of the layer (\eg, Attention and MLP) and $\mathcal{D}$ is the distance metric like L2 distance, \ie, $\mathcal{D} (F(\mathbf{x}), \mathbf{x}) = \|F(\mathbf{x}) - \mathbf{x}\|_2$.
In other words, the representation shift reflects the extent to which each token is emphasized by the function. $F$.\\
Our central hypothesis is that critical tokens tend to have a higher representation shift, as the network encourages them to emphasize the core information or suppresses redundant signals.
Conversely, the tokens with minimal representation shift are likely to be irrelevant to target tasks.
\input{figure/fig_qa}\\
\noindent To validate this hypothesis, we conduct toy experiments with DeiT-S~\cite{touvron2021training} on image classification (ImageNet1K~\cite{deng2009imagenet}) and UMT-B~\cite{li2023unmasked} on video-text retrieval (MSRVTT~\cite{xu2016msr}), and summarize the results in~\Cref{fig:toy}.
For comparison, we first evaluated token importance with attention-based metrics and our representation shift, respectively, and then dropped the tokens having the lowest $k$ scores at each layer ([0,2,4,6,8]). 
We use $k=40$ for DeiT and $k=1100$ for UMT.
For attention-based scoring, we opt \Cref{eq:cls} for DeiT used in \cite{liang2022not, long2023beyond}, and \Cref{eq:sum} for UMT since the class token is generally absent in video transformers.
Also, for representation shift, we compute the L2 distance between the representation of the tokens before and after the attention layer as $\Delta \mathbf{x} = \|\text{SA}(\text{LN}(\mathbf{x})) - \mathbf{x}\|_2 \in \mathbb{R}^{L}$.
As summarized in~\Cref{fig:toy}, pruning based on representation shift achieves competitive or better performance compared to pruning with prevalent attention-based scores.
We demonstrate that the representation shift is a sufficient approximation of the token importance as well as conventional attention-based scores.
Notably, our method introduces no additional learnable parameters and remains applicable even when intermediate attention maps are inaccessible, as in the case of FlashAttention.
\\
We also conduct qualitative analysis of the representation shift (\Cref{fig:qa}) in the middle of DeiT. Interestingly, it captures the foreground object, which aligns with the concept of saliency detection. In other words, we can suppress the noise from the tokens irrelevant to the main content by compressing them based on the proposed scores.
Based on quantitative and qualitative analysis, we underscore the effectiveness of the representation shift for token importance.
In the following subsection, we will provide a thorough investigation of the representation shift.

\input{figure/fig_toy2}
\input{table/retrieval}
\subsection{Exploration on representation shift}
\noindent\textbf{Operation choice.}
Given $\mathbf{x} \in \mathbb{R}^{L\times C}$, the attention blocks of Vision Transformers are typically computed as
\begin{align}
    &\mathbf{x}^\prime= \text{SA}(\text{LN}(\mathbf{x})) + \mathbf{x}, \label{eq:att}\\ 
    &\hat{\mathbf{x}} = \text{MLP}(\text{LN}(\mathbf{x}^\prime)) + \mathbf{x}^\prime, \label{eq:MLP}
\end{align}
where LN is Layer Normalization.
We investigate the impact of the operation choice for representation shift, especially for three cases: representation shift through (i) attention as $\Delta \mathbf{x} = \mathcal{D} (\text{SA}(\text{LN}(\mathbf{x})),\mathbf{x})$, (ii) MLP as $\Delta \mathbf{x} = \mathcal{D} (\text{MLP}(\text{LN}(\mathbf{x^\prime})), \mathbf{x^\prime})$, and (iii) entire attention block including~\Cref{eq:att,eq:MLP} as $\Delta \mathbf{x} = \mathcal{D} (\hat{\mathbf{x}},\mathbf{x})$.
We conducted ablation experiments to evaluate the efficacy of each metric as alternatives for token importance. 
Under the same settings of the previous section, we prune a fixed number of tokens per layer based on the computed L2 distance scores and evaluate the impact on overall model performance.
\Cref{fig:toy21} reveals that token pruning guided by the representation shift through sole MLP generally outperforms other metrics across the layer and models.
Since the attention layer inherently facilitates information exchange across tokens, its transformation may be more diffuse. 
In contrast, the MLP operates on each token independently, leading to a more discriminative representation shift that captures token-specific contributions.
Based on these findings, we adopt the representation shift at MLP as our primary measure for token importance.\\
\noindent\textbf{Distance metrics.}
We further explore which distance metric $\mathcal{D}$ is most appropriate for estimating the representation shift.
A straightforward approach is the (i) \textbf{L2} norm as $\mathcal{D}(\mathbf{x},\mathbf{y}) = \|\mathbf{x}-\mathbf{y}\|_2$, which computes the Euclidean distance between input and output representations, capturing the absolute magnitude of the transformation.
We also study the efficacy of (ii) \textbf{L1} Norm as $\mathcal{D}(\mathbf{x},\mathbf{y}) = \|\mathbf{x}-\mathbf{y}\|_1$, which is more robust to the outliers.
Additionally, (iii) cosine distance (\textbf{Cos}), \ie, $(\mathcal{D}(\mathbf{x},\mathbf{y}))_i = 1 - \frac{\mathbf{x}_i \cdot \mathbf{y}_i}{\|\mathbf{x}_i\|\|\mathbf{y}_i||}$, computes angular difference between vectors, emphasizing directional change rather than magnitude.
For comparison of distance metrics, we compute the representation shift before and after the MLP layer as $\mathcal{D} (\text{MLP}(\text{LN}(\mathbf{x^\prime})), \mathbf{x^\prime})$, and drop the tokens.
As shown in~\Cref{fig:toy22}, the L2 distance consistently produces more robust results as a token importance compared to other distance metrics. 
Our analysis indicates that cosine similarity is suboptimal for assessing token importance in the deeper layers of Transformers. 
Also, although the L1 distance performs favorably at the first layer, it consistently underperforms relative to the L2 distance in subsequent layers.
Therefore, we will use L2 distance for representation shift as the default distance metric.

%% file: figure/fig_qa.tex
\begin{figure}[t!]
\centering
\includegraphics[width=0.48\columnwidth]{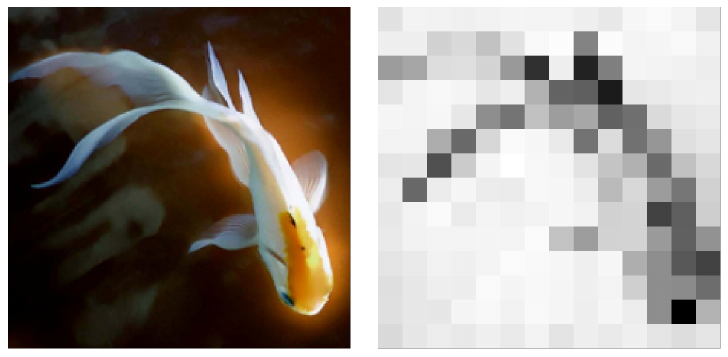}
\hfill
\includegraphics[width=0.48\columnwidth]{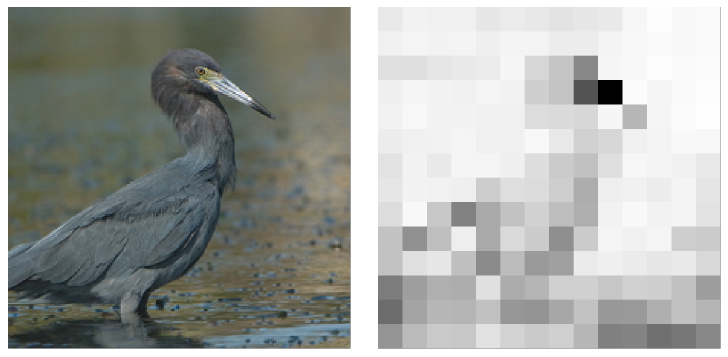}
\includegraphics[width=0.48\columnwidth]{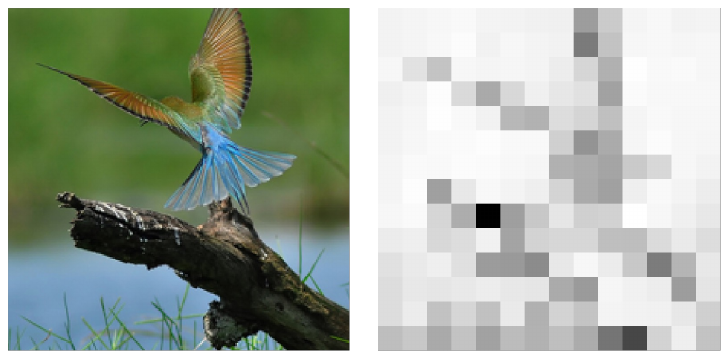}
\hfill
\includegraphics[width=0.48\columnwidth]{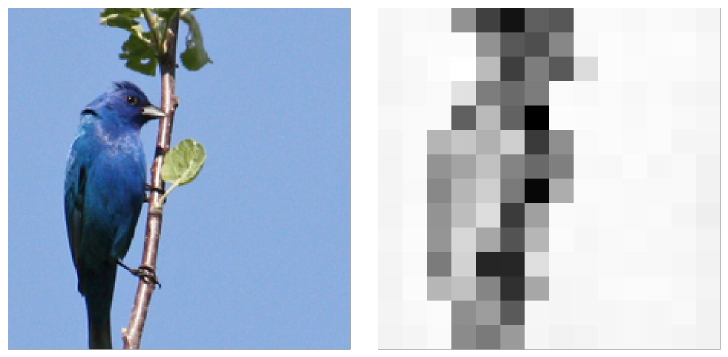}
\vspace{-5pt}
\caption{Visualization of representation shift. Given the image (left), we visualize (right) the representation shift ($\Delta\mathbf {x}$) of each token before and after the attention layer.}
\label{fig:qa}
\vspace{-15pt}
\end{figure}

%% file: figure/fig_toy2.tex
  \begin{figure}[t]
    \centering
    \begin{subfigure}[b]{\columnwidth}
      \centering
      \includegraphics[width=0.48\textwidth]{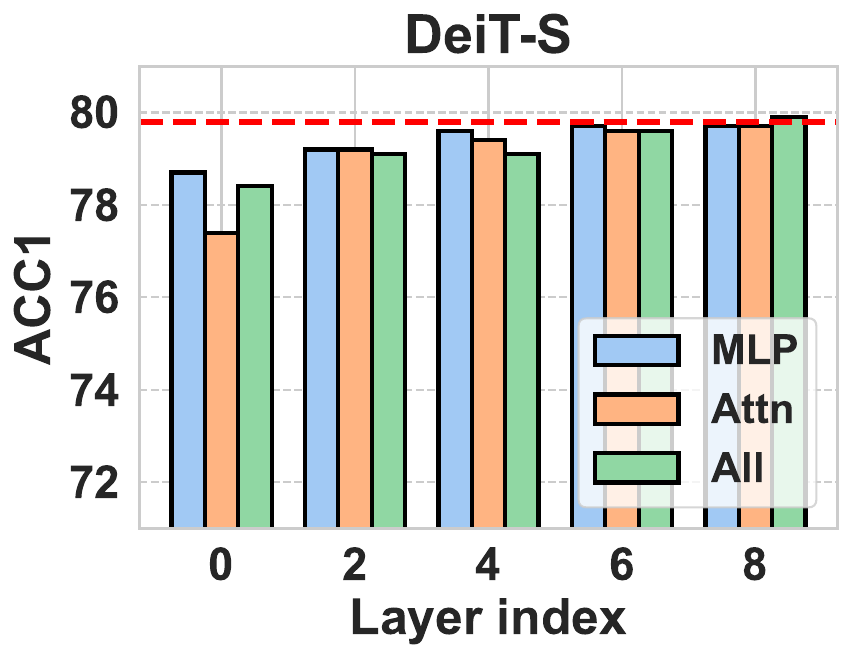}
      \hfill
      \includegraphics[width=0.48\textwidth]{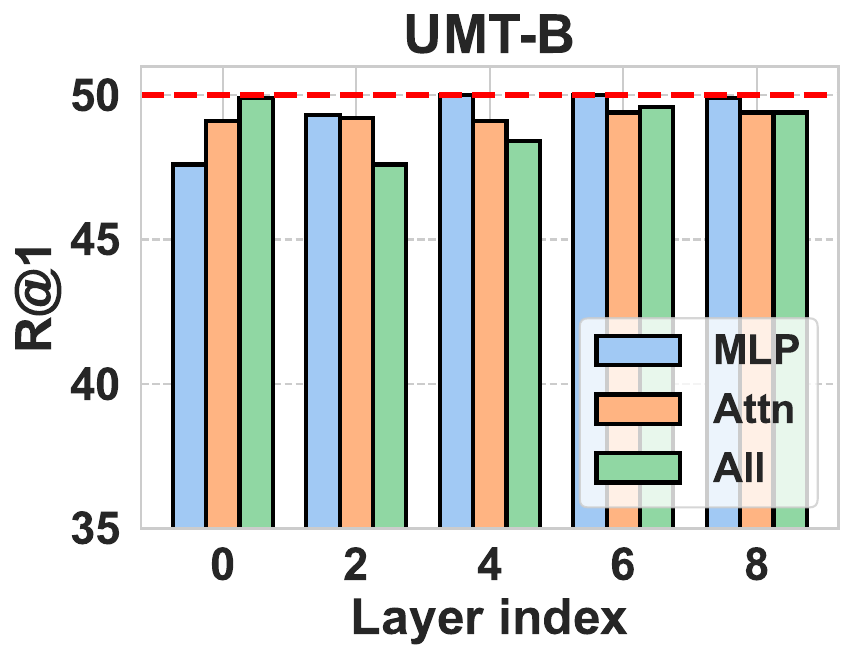}
      \vspace{-5pt}
      \caption{Operation choice}
      \label{fig:toy21}
    \end{subfigure}
    \begin{subfigure}[b]{\columnwidth}
      \centering
      \includegraphics[width=0.48\textwidth]{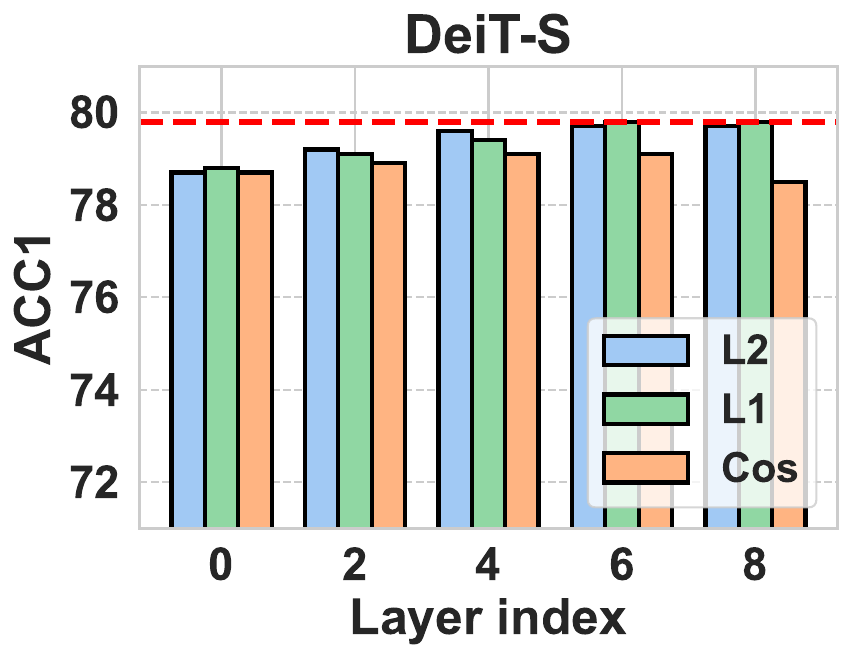}
     \hfill
      \includegraphics[width=0.48\textwidth]{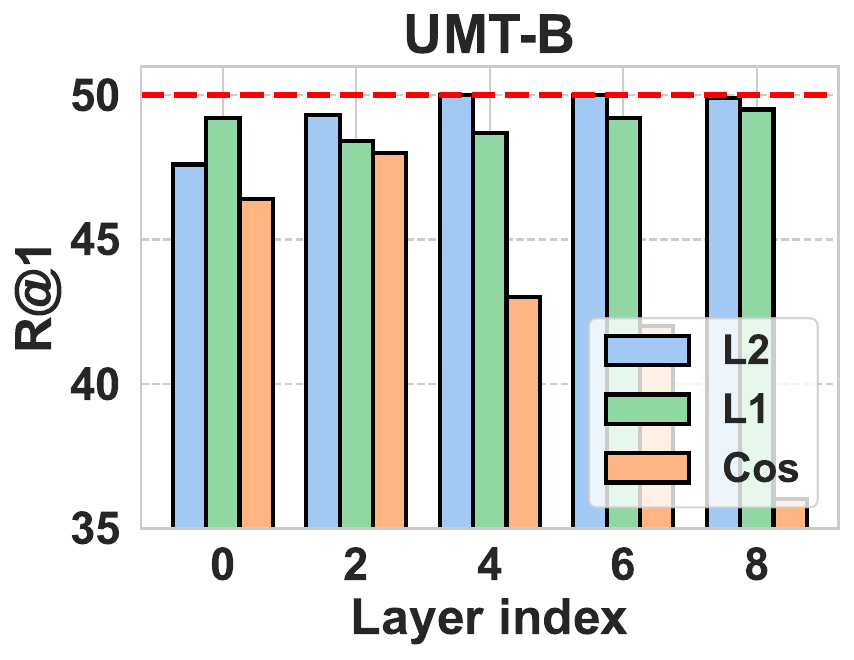}
      \vspace{-5pt}
      \caption{Distance metric}
      \label{fig:toy22}
    \end{subfigure}
    \vspace{-15pt}
    \caption{Analysis on (a) operation choice and (b) distance metric for representation shift. In our experiments, we evaluate the impact of operation choice by pruning tokens based on the representation shift computed using the L2 norm for each candidate operation.
    Similarly, for the analysis of distance metric selection, we prune tokens using each distance metric with the MLP layer.}
    \label{fig:toy2}
\vspace{-10pt}
  \end{figure}

%% file: table/retrieval.tex
\begin{table*}[t]
\small
\centering
\begin{tabular}{c|l|c|cc|c|cc}
\toprule
        \multicolumn{1}{c|}{\multirow{4}{*}{Dataset}} & \multicolumn{1}{c|}{\multirow{2}{*}{Metric}} & \multicolumn{3}{c|}{UMT-B~\cite{li2023unmasked}} & \multicolumn{3}{c}{UMT-L~\cite{li2023unmasked}}    \\ 
        & \multicolumn{1}{c|}{} & \multicolumn{1}{c}{Base} & \multicolumn{1}{c}{Attn} & \multicolumn{1}{c|}{Ours} & \multicolumn{1}{c}{Base} & \multicolumn{1}{c}{Attn} & \multicolumn{1}{c}{Ours} \\  \cline{2-8} 
                                 & Throughput $\uparrow$ &  32   &     57 ($\times$1.78)  &    \textbf{175} ($\times$5.47) &     12  &     23 ($\times$1.91) &  \textbf{66} ($\times$5.50) \\ 
                                & GFlops $\downarrow$   &  303.3            &  156.4  &  156.4  &  984.6 &  478.5  &  478.5  \\ 
\midrule
\multirow{3}{*}{MSRVTT~\cite{xu2016msr}}                
                                & R@1 $\uparrow$        &   50.0  &   47.6  &   \textbf{48.0}   &   58.7  &   50.2  &  \textbf{56.5} \\
                                & R@5 $\uparrow$        &   76.8  &   74.1  &   \textbf{74.4}   &  81.3   &   72.7  &  \textbf{79.6}  \\ 
                                & R@10 $\uparrow$       &   83.9  &   81.7  &   \textbf{82.0}   &   86.8  &   80.3  &  \textbf{86.0}  \\ \hline
\multirow{3}{*}{MSVD~\cite{chen2011collecting}}           
                                & R@1 $\uparrow$        &   62.1  &   \textbf{60.3}  &   57.7  &   70.3  &   64.0  &   \textbf{67.9}  \\
                                & R@5 $\uparrow$        &   89.3  &   \textbf{83.7}  &   80.5  &   89.3  &   84.4  &   \textbf{87.5}  \\ 
                                & R@10 $\uparrow$       &   93.2  &   \textbf{89.0}  &   86.4  &   93.2  &   89.7  &   \textbf{92.2}  \\\hline
\multirow{3}{*}{ActivityNet~\cite{caba2015activitynet}}    
                                & R@1 $\uparrow$        &   57.2  &   \textbf{54.2}  &   50.3  &   65.6  &   53.2  &   \textbf{62.9}  \\
                                & R@5 $\uparrow$        &   83.7  &   \textbf{81.1}  &   78.5  &   89.1  &   80.3  &   \textbf{87.3}  \\ 
                                & R@10 $\uparrow$       &   91.6  &   \textbf{89.6}  &   88.1  &   94.9  &   88.8  &   \textbf{93.8}  \\\hline
\multirow{3}{*}{DiDeMo~\cite{anne2017localizing}}         
                                & R@1 $\uparrow$        &   62.1  &   \textbf{57.7}  &   56.9  &   70.8  &   58.2  &   \textbf{67.3}  \\
                                & R@5 $\uparrow$        &   86.8  &   82.7  &   \textbf{83.3}  &   90.6  &   83.8  &   \textbf{89.1} \\ 
                                & R@10 $\uparrow$       &   92.1  &   88.6  &   \textbf{89.2}  &   94.5  &   89.9  &   \textbf{93.1}  \\ \hline
\multirow{3}{*}{LSMDC~\cite{rohrbach2017movie}}          
                                & R@1 $\uparrow$        &   32.7  &   29.0  &   \textbf{30.0}  &   42.2  &   34.4  &   \textbf{39.8}  \\
                                & R@5 $\uparrow$        &   54.1  &   50.1  &   \textbf{51.1}  &   64.9  &   56.6  &   \textbf{62.9}  \\ 
                                & R@10 $\uparrow$       &   63.3  &   59.2  &   \textbf{59.7}  &   72.3  &   64.1  &   \textbf{70.0}  \\\hline
\multirow{3}{*}{SSV2-label~\cite{lei2022revealing}}     
                                & R@1 $\uparrow$        &   64.0  &   58.0  &   \textbf{59.1}  &   72.4  &   60.6  &   \textbf{69.3}  \\
                                & R@5 $\uparrow$        &   88.3  &   83.9  &   \textbf{84.4}  &   93.4  &   85.7  &   \textbf{91.0}  \\ 
                                & R@10 $\uparrow$       &   92.9  &   \textbf{90.8}  &   90.7  &   96.7  &   91.1  &   \textbf{94.9}  \\\hline
\multirow{3}{*}{SSV2-Template~\cite{lei2022revealing}}  
                                & R@1 $\uparrow$        &   74.6  &   65.4  &   \textbf{69.0}  &   78.4  &   67.5  &   \textbf{74.8}  \\
                                & R@5 $\uparrow$        &   93.9  &   91.3  &   \textbf{92.4}  &   95.9  &   91.9  &   \textbf{95.0}  \\
                                & R@10 $\uparrow$       &   96.8  &   95.0  &   \textbf{95.3}  &   97.8  &   94.9  &   \textbf{97.5}  \\ \hline  
\end{tabular}
\vspace{-5pt}
\caption{
     Video-text retrieval on MSRVTT~\cite{xu2016msr}, MSVD~\cite{chen2011collecting}, ActivityNet~\cite{caba2015activitynet}, DiDeMo~\cite{anne2017localizing}, LSMDC~\cite{rohrbach2017movie}, SSV2-Label/Template~\cite{lei2022revealing}.
}
\label{table:retrieval}
\vspace{-15pt}
\end{table*}

%% file: sec/4_exp.tex
\section{Experiments}
In this section, we will present the results of video understanding tasks in~\Cref{sec:4.1}, image classification in~\Cref{sec:4.2}, and analysis of the proposed method in \Cref{sec:4.3}.
\subsection{Video Understandings}
\label{sec:4.1}
\noindent\textbf{Settings.} 
To validate the efficacy of the representation shift, we first conducted token pruning based on representation shift with several video tasks, where the large number of tokens across frames imposes significant computational costs.
We use the UMT~\cite{li2023unmasked}, a Video Transformer built with vanilla attention, as our baseline for video-text retrieval~\cite{xu2016msr,chen2011collecting,caba2015activitynet,anne2017localizing,rohrbach2017movie,lei2022revealing}, and video question-answering~\cite{xu2017video}.
For comparison with attention-based scores, we also use the averaged attention scores as in~\Cref{eq:sum}, since the class token is not available at Video Transformer. 
We progressively reduce the number of tokens by 20\% and 10\% in each of the first three layers of UMT for video-text retrieval and video question-answering, respectively, by applying pruning based on both metrics. 
FlashAttention is used in the case of representation shift, as the attention-based score is not compatible with it.
All experiments are conducted in a training-free manner.

\input{table/retrieval_vid-tldr}
\input{table/qa}

\noindent\textbf{Video-text retrieval.} 
In video-text retrieval, the model retrieves the most related text given a video (video-to-text retrieval, V2T) or finds the most relevant video for a text query (text-to-video retrieval, T2V). 
We report the harmonic mean of results of V2T and T2V on seven benchmarks: MSRVTT~\cite{xu2016msr}, MSVD~\cite{chen2011collecting}, ActivityNet~\cite{caba2015activitynet}, DiDeMo~\cite{anne2017localizing}, LSMDC~\cite{rohrbach2017movie}, SSV2-Label/Template~\cite{lei2022revealing}.
For comparing the efficiency, we also measure and provide both FLOPs (G) and throughput (vid/s) using a single NVIDIA RTX A6000 with a batch size of 20, given the video consisting of 12 frames with $224^2$ resolutions.
Given the baseline model without token pruning (\textbf{Base}), we apply token pruning with attention-based scores (\textbf{Att}) and representation shift (\textbf{Ours}), respectively.
The results are presented in~\Cref{table:retrieval}.
Since our representation shift enables the token pruning to work with FlashAttention, it brings a promising $5.47\times$ and $5.5\times$  speed-up in UMT-B and UMT-L, respectively.
Our approach nearly doubles the throughput compared to token pruning methods based on traditional attention scores with standard attention.
Further, despite the faster inference, our approach has shown competitive or even better performance, achieving up to 9.7\% R@1 gain, especially with UMT-L on ActivityNet compared to attention-based pruning.
On average, we observe a +7.2\% improvement in R@1 with UMT-L.
It is worth noting that applying token pruning with representation shift offers a more favorable speed-accuracy trade-off than simply downscaling the model, as UMT-L with representation shift (66 vid/s) achieves approximately 2$\times$ higher throughput than base UMT-B (32 vid/s), while consistently surpassing it.\\
We further explore the applicability of representation shift with other token compression work by replacing the importance metric of vid-TLDR~\cite{choi2024vid}, a token merging method for efficient video transformer.
Following the original configuration of vid-TLDR, including the reduction ratio and layer choice, we report the results on video-text retrieval.
In~\Cref{table:vidtldr}, we demonstrate the solid advantage of representation shift with other token compression.
Originally, vid-TLDR employed an attention-based metric to detect salient regions of the image, which was thus incompatible with FlashAttention.
However, by substituting the importance metric with our representation shift, we can harness the efficiency of FlashAttention along with vid-TLDR.
Specifically, under the same reduction ratio, our representation shift achieves an average speed-up of 3.74$\times$ and 3.67$\times$ in UMT-B and UMT-L with the minimal performance drop.\\
\noindent\textbf{Video question-answering.} 
We also demonstrate the efficiency of the proposed approach in video question-answering (video QA) tasks.
In video QA, the model generates responses to questions related to a given video.
To evaluate this, we assess each method on MSRVTT-QA, MSVD-QA benchmarks~\cite{xu2017video}, summarizing the results in~\Cref{table:qa}.
Similar to video-text retrieval, we compare the three cases: the baseline model without pruning (\textbf{Base}), the model with attention-based token pruning (\textbf{Att}), and (\textbf{Ours}).
Compared to the Base model, we demonstrate a promising improvement, achieving approximately 4$\times$/3.83$\times$ higher throughput in UMT-B/L.
Further, despite being faster than conventional attention-based pruning, our approach achieves comparable or even better performance.
Notably, in the UMT-L, we observe significant improvements of 0.5\% and 0.7\% on MSRVTT and MSVD, respectively. 

\subsection{Image Classification}
\label{sec:4.2}

\input{table/deit}
\noindent\textbf{Vision Transformers.}
We experiment on image classification with ImageNet1K~\cite{deng2009imagenet}.
For vision transformers, we use DeiT~\cite{touvron2021training} without additional training, and report the top-1 accuracy and throughput with a batch size of 512.
For comparison, we use attention scores for class token (\Cref{eq:cls}) used in EViT~\cite{liang2022not}, and BAT~\cite{long2023beyond}.
For the representation shift, we use the same settings (L2, MLP) as video understandings, along with FlashAttention.
After quantifying the importance of the tokens in the [1,4,7] layers of DeiT, we pruned the 20\% tokens at each layer.
As shown in~\Cref{table:deit}, although the same proportion of tokens is pruned, our method consistently outperforms the attention-based scores.
Specifically, combined with FlashAttention, the representation shift achieves $1.2\times$ higher throughput with the gain of  +2.8\%, +5.7\%, and +2.7\% accuracy gain in DeiT-T/S/B compared to attention-based scoring.
We believe that representation shift provides more robust importance scores than traditional attention scores, resulting in a significant performance gap.

\input{figure/fig_qa2}

\noindent\textbf{CNN and SSM.}
Since representation shift is a model-agnostic approach to estimate the token importance, it naturally extends to other architectures, which have been under-explored in previous token compressions.
For this, we first conduct experiments with ResNet~\cite{he2016deep} on ImageNet1K.
In CNNs, we measure the representation shift before and after each stage, as ResNet does not contain MLPs.
Since the convolutional operation in ResNet only works with a 2D grid structure, token pruning in CNNs cannot be performed in a straightforward manner.
So, we consider two variants of token pruning: i) removing the least important tokens from each row and column (Token-wise, \textbf{T-W}), and ii) averaging the representation shift across each row and column and then pruning tokens line by line from those rows and columns with the lowest average values (Line-wise, \textbf{L-W}), akin to~\cite{su2024removing}.
Specifically, by each approach, we remove 8 columns and 8 rows after the first stage, and 4 columns and 4 rows after the second stage.
As the resolutions are changed after token compression in CNNs, we finetune the model for 100 epochs, including 10 cooldown epochs to refine this change.
\Cref{table:resnet} reveals that both pruning approaches with representation shift bring substantial throughput improvements in ResNet.
We observe at least 18\% speed up in both pruning approaches.
Especially, line-wise pruning shows very competitive performance with the base ResNet without pruning, achieving the higher throughput of 7112/3553 (img/s) compared to the original throughput of 5811/2927 (img/s) in ResNet-34/50.
\input{table/resnet}

\input{table/vim}

\noindent We also validate representation shift with State Space Model (SSM) using Vision Mamba (ViM)~\cite{zhu2024vision} in \Cref{table:vim}.
Overall, we largely follow the settings of ToP-ViM~\cite{zhan2024exploring}, which is designed for accelerating SSM by pruning tokens based on the activated values.
We observe the improvements of +0.4\% on ViM-T under a similar throughput of Top-ViM.
These results suggest that representation shift is a generalizable approach for various architectures.





\subsection{Analyses}
\label{sec:4.3}
\noindent\textbf{Qualitative Results.}
For a deeper understanding of the behavior of representation shift, we provide a qualitative analysis with a visualization.
In~\Cref{fig:qa2}, given the image sample (left), we qualitatively compare the attention-based scores of~\Cref{eq:cls} used in~\cite{liang2022not, long2023beyond}, and our proposed representation shift using the DeiT-B~\cite{touvron2021training} consisting of 12 attention layers.
To investigate the behavior of each method across early, middle, and deeper layers, we evaluate them at the 1st, 5th, and 9th layers of the model.
First, in the early stage (L=1), the attention map generally shows low reliability as discussed in prior works~\cite{choi2024vid,liang2022not}, which is not a desirable option for token importance.
On the other hand, our representation shift successfully detects the foreground object even in the first layer.
In the middle layer (L=5), representation shift still captures the main content better than attention scores.
Lastly, in Vision Transformers, it is well-known that global information is gathered in a few specific tokens as the layer passes~\cite{darcet2023vision}, having higher attention scores.
In this respect, it would be better to mimic the attention map in the latter layer (L=9) to avoid information loss for retaining the informative tokens.
To summarize, the representation shift mitigates the low reliance of attention scores in the early layer and finds the salient region till the middle layer, helping the model to capture fine-grained patterns.
Further, it enables capturing the token having high-level semantics in the latter layer.


\input{figure/fig_qa3}
\noindent Additionally, we visualize the representation shift through each stage of ResNet-50~\cite{he2016deep} in~\Cref{fig:qa3}.
The results reveal that the embedding of the foreground tokens tends to have a more drastic shift than background tokens in every stage.
In other words, the network updates the foreground tokens more aggressively, while background tokens, being less critical, undergo only subtle updates. 
Consequently, the representation shift inherently serves as informativeness of the token to the task, allowing for token pruning without compromising overall performance as shown in~\Cref{table:resnet}.

\noindent\textbf{Reliability analysis.} 
To assess the reliability of representation shift as an importance metric, we conduct an extreme pruning experiment using DeiT-S~\cite{touvron2021training} on ImageNet-1K~\cite{deng2009imagenet}, where we retain either the top or the bottom 50\% of tokens ranked by their representation shift scores.
As shown in~\Cref{tab:rel}, across all transformer layers (L1–L11), retaining the top 50\% consistently yields significantly higher accuracy than keeping the bottom 50\%, demonstrating the robustness of the importance signal.
On average, the top 50\% selection achieves 78.0\% accuracy, whereas the bottom 50\% only reaches 51.7\%, resulting in a substantial performance gap of 26.3\%. 
This consistent gap across layers validates that representation shift effectively identifies informative tokens, supporting its reliability.

\input{table/reliability}



%% file: table/retrieval_vid-tldr.tex
\begin{table*}[t]
\centering
\small
\begin{tabular}{c|l|ccc|ccc}
\toprule

\multicolumn{1}{c|}{\multirow{2}{*}{Dataset}} & \multicolumn{1}{c|}{\multirow{2}{*}{Metric}} & \multicolumn{3}{c|}{UMT-B~\cite{li2023unmasked}} & \multicolumn{3}{c}{UMT-L~\cite{li2023unmasked}} \\ 

\multicolumn{1}{c|}{} & \multicolumn{1}{c|}{} & \multicolumn{1}{c}{Base} & \multicolumn{1}{c}{vid-TLDR} & \multicolumn{1}{c|}{+Ours} & \multicolumn{1}{c}{Base} & \multicolumn{1}{c}{vid-TLDR} & \multicolumn{1}{c}{+Ours} \\

\midrule
\multirow{2}{*}{MSRVTT~\cite{xu2016msr}}
    & Throughput $\uparrow$     &  32.0  &  43.6 ($\times$1.36)  &  \textbf{131.4} ($\times$4.10)  &  12.0  &  19.1 ($\times$1.59)  &  \textbf{41.0} ($\times$3.42)  \\
    & R@1 $\uparrow$            &  50.0  &  \textbf{50.8}  &  \textbf{50.8}  &  \textbf{58.7}  &  58.5  &  58.6  \\ \midrule
\multirow{2}{*}{MSVD~\cite{chen2011collecting}}
    & Throughput $\uparrow$     &  32.0  &  42.2 ($\times$1.32)  &  \textbf{131.1} ($\times$4.10)  &  12.0  &  19.9 ($\times$1.66)  &  \textbf{40.0} ($\times$3.33)  \\
    & R@1 $\uparrow$            &  62.1  &  62.7  &  \textbf{62.8}  &  70.3  &  70.4  &  \textbf{70.6}  \\ \midrule
\multirow{2}{*}{ActivityNet~\cite{caba2015activitynet}}
    & Throughput $\uparrow$     &  32.0  &  34.2 ($\times$1.07)  &  \textbf{114.7} ($\times$3.58)  &  12.0  &  18.1 ($\times$1.51)  &  \textbf{40.4} ($\times$3.34)  \\
    & R@1 $\uparrow$            &  \textbf{57.2}  &  56.6  &  56.8  &  65.6  &  65.2  &  \textbf{66.0}  \\ \midrule
\multirow{2}{*}{DiDeMo~\cite{anne2017localizing}}
    & Throughput $\uparrow$     &  32.0  &  33.9 ($\times$1.06)  &  \textbf{129.3} ($\times$4.05)  &  12.0  &  18.3 ($\times$1.53)  &  \textbf{51.5} ($\times$4.29)  \\
    & R@1 $\uparrow$            &  62.1  &  \textbf{62.4}  &  62.0  &  70.8  &  70.4  &  \textbf{70.9}  \\ \midrule
\multirow{2}{*}{LSMDC~\cite{rohrbach2017movie}}
    & Throughput $\uparrow$     &  32.0  &  36.5 ($\times$1.14)  &  \textbf{110.7} ($\times$3.46)  &  12.0  &  16.6 ($\times$1.38)  &  \textbf{50.8} ($\times$4.23)  \\
    & R@1 $\uparrow$            &  \textbf{32.7}  &  32.4  &  32.4  &  \textbf{42.2}  &  41.9  &  41.9  \\ \midrule
\multirow{2}{*}{SSV2-label~\cite{lei2022revealing}}
    & Throughput $\uparrow$     &  32.0  &  34.1 ($\times$1.07)  &  \textbf{114.4} ($\times$3.58)  &  12.0  &  16.4 ($\times$1.37)  &  \textbf{39.9} ($\times$3.33)  \\
    & R@1 $\uparrow$            &  \textbf{64.0}  &  63.8  &  63.5  &  \textbf{72.4}  &  72.1  &  71.8  \\ \midrule
\multirow{2}{*}{SSV2-Template~\cite{lei2022revealing}}
    & Throughput $\uparrow$     &  32.0  &  38.1 ($\times$1.19)  &  \textbf{106.6} ($\times$3.33)  &  12.0  &  16.0 ($\times$1.33)  &  \textbf{45.2} ($\times$3.77)  \\
    & R@1 $\uparrow$            &  \textbf{74.6}  &  74.0  &  73.9  &  78.4  &  78.1  &  \textbf{78.5}  \\

\bottomrule

\end{tabular}
\vspace{-7pt}
\caption{
    Extensibility of representation shift with other token compression, vid-TLDR~\cite{choi2024vid}.
    +Ours indicates the results of vid-TLDR ~\cite{choi2024vid} after replacing the importance metric with representation shift and adopting FlashAttention.
}
\label{table:vidtldr}
\vspace{-18pt}
\end{table*}

%% file: table/qa.tex
\begin{table}[t]
\centering
\small
\setlength{\tabcolsep}{3pt}
\begin{tabular}{l|c|c|c|c}  
\toprule
\multirow{2}{*}{\textbf{Method}} 
& \multirow{2}{*}{\textbf{GFlops}} 
& \multirow{2}{*}{\textbf{Throughput}} 
& \multirow{2}{*}{\textbf{MSR-QA}} 
& \multirow{2}{*}{\textbf{MSVD-QA}} \\
& & & & \\ 
\midrule
UMT-B~\cite{li2023unmasked}   & 303.3  &   32 & 44.9 & 48.1  \\
UMT-B-Attn                    & 217.7  &   39(x1.22) & 44.8 & 46.5 \\
UMT-B-Ours                    & 217.7  &  128(x4.00) & 44.6 & 47.0 \\ \midrule
UMT-L~\cite{li2023unmasked}   & 984.6  &   12 & 49.5 & 55.2 \\
UMT-L-Attn                    & 690.5  &   15(x1.25) & 49.5 & 54.2 \\
UMT-L-Ours                    & 690.5  &   46(x3.83) & 49.0 & 54.9 \\
\bottomrule
\end{tabular}
\vspace{-5pt}
\caption{
   Video question-answering on MSRVTT-QA~\cite{xu2017video} \& MSVD-QA~\cite{xu2017video}.
}
\label{table:qa}
\vspace{-20pt}
\end{table}

%% file: table/deit.tex
\begin{table}[t]
\centering
\small
\begin{tabular}{c|c|c|cc}
\toprule
Method & Metric & Base & Attn & Ours\\
\midrule
\multirow{3}{*}{Deit-T} & Acc & 72.1 &65.5 & 68.3 \\
& Throughput & 6725 & 10949 & 13296\\
& GFLOPs & 1.3 & 0.8 & 0.8 \\
\midrule
\multirow{3}{*}{Deit-S} & Acc & 79.8 & 72.1 & 77.8 \\
& Throughput & 3002 & 4844 & 5948\\
& GFLOPs & 4.6 & 3.0 & 3.0 \\
\midrule
\multirow{3}{*}{Deit-B} & Acc & 81.8 & 76.9 & 79.6\\
& Throughput & 1037 & 2065 & 2428\\
& GFLOPs & 17.6 & 11.5 & 11.5 \\
\bottomrule
\end{tabular}
\vspace{-5pt}
\caption{ImageNet1K~\cite{deng2009imagenet} classification results with DeiT~\cite{touvron2021training}.}
\label{table:deit}
\vspace{-15pt}
\end{table}

%% file: figure/fig_qa2.tex
\begin{figure*}[t!]
\centering
\includegraphics[trim=0 0 0 0,clip, width=0.9\linewidth]{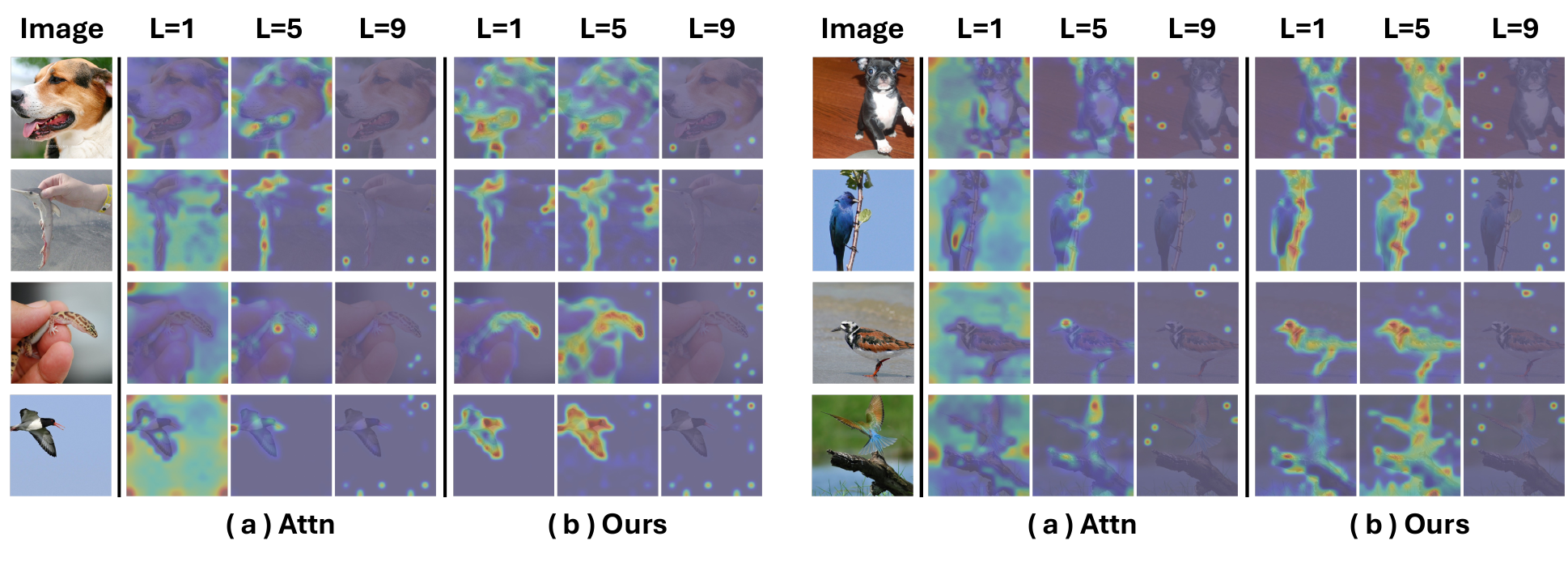}
 \vspace{-15pt}
 \caption{Qualitative comparison between attention scores (Attn) and representation shift (Ours). Given each sample, we visualize (a) the attention scores with respect to the class token and (b) representation shift in the [1,5,9] layers of the DeiT-B~\cite{touvron2021training}.}
\label{fig:qa2}
\vspace{-15pt}
\end{figure*}

%% file: table/resnet.tex
\begin{table}[t]
\centering
\small
\begin{tabular}{c|c|c|cc}
\toprule
Method & Metric & Base & L-W & T-W\\
\midrule
\multirow{3}{*}{ResNet-34} & Acc & 73.2 & 72.8 & 72.2 \\
& Throughput & 5811 & 7112 & 6867\\
& GFLOPs & 3.7 & 2.5 & 2.5 \\
\midrule
\multirow{3}{*}{ResNet-50} & Acc & 76.1 & 76.4 & 75.9 \\
& Throughput & 2927 & 3553 & 3489 \\
& GFLOPs & 4.1 & 2.7 & 2.7 \\
\bottomrule
\end{tabular}
\vspace{-5pt}
\caption{ImageNet1K~\cite{deng2009imagenet} classification results with ResNet~\cite{he2016deep}. L-W: Line-wise pruning, T-W: Token-wise pruning}
\label{table:resnet}
\vspace{-10pt}
\end{table}

%% file: table/vim.tex
\begin{table}[t]
\centering
\small
\begin{tabular}{c|c|c|cc}
\toprule
Method & Metric & Base & ToP-ViM~\cite{zhan2024exploring} & Ours\\
\midrule
\multirow{3}{*}{ViM-T} & Acc & 76.1 & 75.1 & 75.5 \\
& Throughput & 1603 & 1758 & 1754\\
& GFLOPs & 1.5 & 1.3 & 1.3 \\
\bottomrule
\end{tabular}
\vspace{-5pt}
\caption{ImageNet1K~\cite{deng2009imagenet} classification results with ViM~\cite{zhu2024vision}.}
\label{table:vim}
\vspace{-15pt}
\end{table}

%% file: figure/fig_qa3.tex
\begin{figure}[t!]
\centering
\includegraphics[trim=0 0 0 0,clip, width=0.9\linewidth]{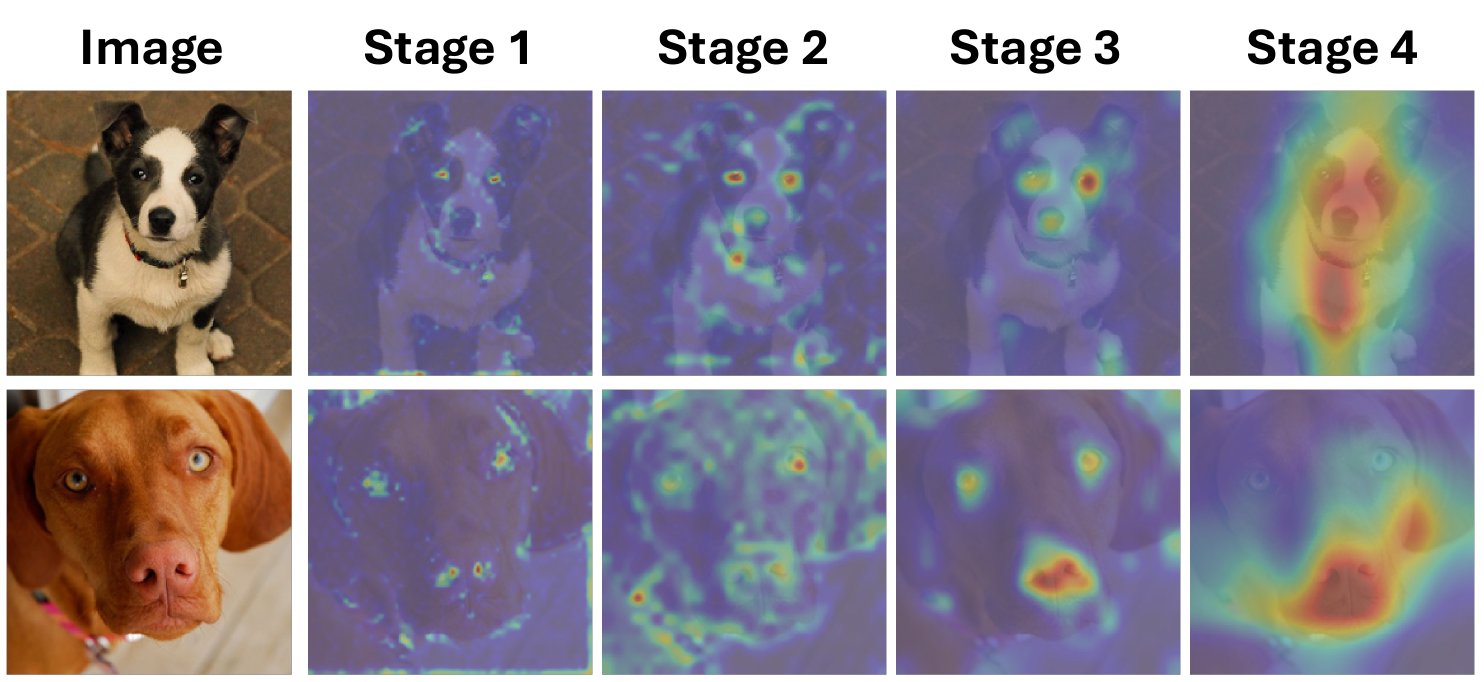}
\caption{Visualization of representation shift in ResNet-50~\cite{he2016deep}.}
\label{fig:qa3}
\vspace{-15pt}
\end{figure}

%% file: table/reliability.tex
\begin{table}[t!]
\small
\centering
  \centering
  \setlength{\tabcolsep}{3pt}
  \begin{tabular}{l|ccccccc}
  \toprule
    Token Selection & L1 & L3 & L5 & L7 & L9 & L11 & Avg \\
    \midrule
    Top 50\%& 76.3 & 76.1 & 78.5 & 79.4 & 78.5 & 78.9 & 78.0 \\
    Bottom 50\% & 51.4 & 51.9 & 47.0 & 49.6 & 56.1 & 54.1 & 51.7 \\
    \bottomrule
  \end{tabular}
\vspace{-5pt}
\caption{Accuracy when retaining top/bottom-50\% tokens}
\label{tab:rel}
\vspace{-15pt}
\end{table}

%% file: sec/5_con.tex
\section{Conclusion}
\label{sec:5}
In this paper, we propose a novel training-free, model-agnostic token importance criterion based on representation shift, which effectively quantifies the information contribution of each operation.
Unlike conventional methods, our approach operates independently of attention maps, allowing seamless integration with FlashAttention while achieving competitive accuracy and substantial inference speed improvements.
Moreover, its applicability extends beyond Transformers to CNNs, making it a versatile approach for enhancing the efficiency of various vision models while preserving performance.
Additionally, we qualitatively demonstrate that our approach successfully detects foreground objects in early and middle layers more effectively than existing methods and informative tokens in latter layers, highlighting its potential as an improved token importance criterion for efficient token compression.

\section*{Acknowledgements.}
This work was partly supported by IITP grant funded by MSIP \& MSIT (No. RS-2024-00443251, No. RS-2024-00457882), NRF grant funded by MSIT (NRF-2023R1A2C2005373), and IITP-ITRC grant funded by MSIT (IITP-2025-RS-2024-00436857).